\documentclass[a4paper]{article}

\usepackage{INTERSPEECH2020}

\usepackage{breqn}
\usepackage{amsthm}
\usepackage{bm}
\usepackage{mathtools}

\usepackage[a4paper, textheight=9.7in, textwidth=6.8in, tmargin=1.5in]{geometry}

\usepackage{algorithm,algpseudocode}
\makeatletter
\let\OldStatex\Statex
\renewcommand{\Statex}[1][3]{%
  \setlength\@tempdima{\algorithmicindent}%
  \OldStatex\hskip\dimexpr#1\@tempdima\relax}
\makeatother

\DeclareMathOperator*{\argmin}{arg\,min}

\newcommand\bos{{\texttt{<s>}\,}}
\newcommand\eos{{\texttt{<\textbackslash{}s>}}}

\usepackage{xcolor}
\usepackage{multirow}
\usepackage{hyperref}

\title{Efficient MDI Adaptation for $n$-gram Language Models}
\name{Ruizhe Huang$^1$, Ke Li$^1$,  Ashish Arora$^1$, Daniel Povey$^2$, Sanjeev Khudanpur$^1$}
\address{
  $^1$Center for Language and Speech Processing, The Johns Hopkins University, Baltimore, USA\\
  $^2$Xiaomi Inc., Beijing, China}
\email{\{huangruizhe09, ashish.arora.88888, dpovey\}@gmail.com, \{kli26, khudanpur\}@jhu.edu}

\begin{document}

\maketitle
% \eightpt
% 
\begin{abstract}
  This paper presents an efficient algorithm for $n$-gram language model adaptation under the minimum discrimination information (MDI) principle, where an out-of-domain language model is adapted to satisfy the constraints of marginal probabilities of the in-domain data. The challenge for MDI language model adaptation is its computational complexity. By taking advantage of the backoff structure of $n$-gram model and the idea of hierarchical training method, originally proposed for maximum entropy (ME) language models \cite{wu2000efficient}, we show that MDI adaptation can be computed in linear-time complexity to the inputs in each iteration. The complexity remains the same as ME models, although MDI is more general than ME. This makes MDI adaptation practical for large corpus and vocabulary. Experimental results confirm the scalability of our algorithm on very large datasets, while MDI adaptation gets slightly worse perplexity but better word error rate results compared to simple linear interpolation.
\end{abstract}

\section{Introduction}

The $n$-gram language model (LM) still plays an important role in today's automatic speech recognition (ASR) pipeline. There are several reasons: (i) $n$-gram LMs can be represented by weighted finite-state transducers (WFST) and integrated into first-pass decoding \cite{mohri2002weighted}, (ii) training and querying $n$-gram LMs are cheaper than neural LMs, (iii) in practice, the best performance is achieved by interpolating $n$-gram and neural LMs \cite{shareghi2019show}.

Consider a common scenario when one hopes to develop an ASR system for a new application, while little training data is available and collecting sufficient domain-specific ({\em in-domain}) data requires a considerable amount of time and efforts. The limited data is too small to estimate a robust LM. Fortunately, we can do better by capitalizing on some large, general-domain background ({\em out-of-domain}) corpus assuming the out-of-domain data may contain much information common with the application domain. This motivates LM adaptation \cite{bellegarda2004statistical, demori1999language}, which is to estimate a robust LM based on both in- and out-of-domain data.

The question is how to combine information from the two sources in a suitable manner?  The commonly used approaches for $n$-gram LMs fall under two categories: model interpolation and constraint-based methods. The model interpolation methods can be either linear (simple linear \cite{JelMer80}, history-dependent \cite{liu2013use}, Bayesian \cite{pusateri2019connecting} interpolation) or non-linear (log-linear \cite{heafield2016normalized} interpolation, fill-up technique \cite{besling1995language}). Note that simple linear interpolation is very effective and probably the most popular adaptation method. Recently, \cite{pusateri2019connecting} found that count-merging, as a special case of maximum a posterior (MAP) adaptation, is theoretically similar to Bayesian interpolation. On the other hand, the constraint-based methods \cite{rosenfeld1996maximum}, such as ME or MDI models, attempt to choose the adapted LM such that it satisfies some constraints in the adaptation domain, while staying as close as possible to some prior distribution, measured by, {\em e.g.}, Kullback-Leibler distance. This paper investigates the MDI adaptation.

There has been previous work on MDI adaptation for $n$-gram LMs \cite{della1992adaptive, rao1995language, rao1997mdi, weintraub1996lm95, reichl1999language, federico1999efficient, kneser1997language, chueh2008reliable, chen2009shrinking, ruiz2012mdi}, with several variants of task definition, {\em e.g.}, adaptation for cache model, within- or cross-corpus adaptation. Although MDI has appealing theoretical properties, the computation is non-trivial and expensive, which grows almost exponentially (detailed in section \ref{sec-algo}) with the size of the vocabulary \cite{demori1999language} in a naive implementation. To reduce the complexity, \cite{kneser1997language, reichl1999language} proposed approximation algorithms and \cite{rosenfeld1994adaptive, Alume2010EfficientEO} devised parallelization to speed up the computation. \cite{federico1999efficient} proposed a linear-time algorithm for unigram constraints. On the other hand, there has been work on ME model that utilizes the back-off structure of the LMs to reduce computational complexity to linear time per iteration \cite{wu2000efficient}, but it is not clear whether the same trick can carry over to MDI which is more general than ME. Besides, regarding model performance, most previous work has found that MDI adaptation performs slightly worse than simple linear interpolation \cite{chen2009shrinking}, but we are interested to see if there can be any difference when operating on very large corpus once we have an efficient MDI algorithm for arbitrary marginal distribution constraints. Moreover, in the experiment, we will propose a novel approach of applying MDI adaptation to improve the first-pass LM while keeping the model size unchanged.

We will review MDI adaptation in section 2. In section 3, we describe our efficient algorithm and describe some implementation concerns. In section 4, we show experiments to demonstrate the scalability of the algorithm and compare the perplexities and word error rates with linear interpolation, the baseline. We will conclude with future work in section 5.

% (Any related work on log-linear model?)

% future work:
% 1. feature selection, history estimation
% 2. KN consistency
% 3. implement regularization in the toolkit

\section{Background}

\subsection{$n$-gram Language Model}

A language model (LM) is a probability distribution over word sequences $W = w_1 w_2 \ldots w_L$, usually reduced to a word-by-word probability via the chain rule $p(W) = \prod_{i=1}^{L} p(w_i | w_{1}^{i})$, where $w_i^j\equiv w_i, w_{i+1}, \ldots, w_j$. An $n$-gram LM assumes that this distribution depends only on the previous $n-1$ words, {\em i.e.}, $p(w_i | w_{1}^{i}) \approx p(w_i | w_{i-n+1}^{i-1})$, where $w_{i-n+1}^{i-1}$ is the history $h_i$ of word $w_i$. We omit the index $i$ when the context is clear. 

Given the vocabulary $V$, $n$-gram LM defines a set of conditional probabilities $p(w|h)$ for any $hw \in V^n$. However, the space $V^n$ is very large that not every $n$-gram $hw$ is seen in the training data, known as the data sparsity problem. Thus, smoothing techniques have been used to estimate the probability $p(w|h)$ for the unseen $n$-grams. The most popular techniques is backing-off. The idea is to recursively estimate $p(w|h)$ of unseen $n$-grams based on the lower order $(n-1)$-gram probabilities $p(w|h')$, where $h' = w_{i-n+2}^{i-1}$, which may have been seen in the corpus. More specifically, 
\begin{equation}
      p(w|h) = \left\{
        \begin{array}{ll}
            p^*(w|h) & hw \textup{ is seen in corpus} \\
            bow(h) \cdot p(w|h')   & \textup{otherwise,} \\
		\end{array} 
		\right. \label{backoff}
\end{equation}
where the discounted probability $p^*(w|h)$ and the back-off weight $bow(h)$ are together to ensure the conditional probability sums to one: $\sum_{w \in V} p(w|h) = 1$. We will consider $n$-gram LMs having back-off structure in the rest of the paper.
In practice, such LMs are stored in ARPA format \cite{slp2}. Note that LMs smoothed by interpolation \cite{slp2} can also be stored as ARPA. We measure the size of an $n$-gram LM as the total number of entries of order $1, \ldots, n$ when the LM is represented as ARPA.

% Note that the model smoothed by interpolation is a special case of back-off model. 

\vspace{-0.1in}
\subsection{MDI Adaptation}

The idea of LM adaptation under the minimum discrimination information (MDI) principle is to compute the adapted distribution such that it satisfies the constraints characterizing in-domain distribution, and also stays closest to the out-of-domain distribution. The constraints are usually expressed as marginal distributions.

% For simplicity, we take trigram LM as an example to illustrate the idea from now on.

Formally, given (i) the vocabulary $V$, (ii) the out-domain LM $p_{out}(w|h)$, (iii) the empirical history distribution $\tilde{p}(h)$ which is commonly approximated by either the in-domain probabilities $p_{in}(h)$ or out-of-domain $p_{out}(h)$, and (iv) K marginal distributions $\tilde{p}(S_i)$ where $S_i \subset V^n, i=1,\ldots, K$ derived from the in-domain data, the adapted LM $p_{ad}(w|h)$ is defined by minimizing the following conditional Kullback-Liebler (KL) divergence:
% \vspace{-0.03in}
\begin{align}
    p_{ad}(w | h) &= \argmin_p D(p || p_{out} | \tilde{p}) \\
    = \argmin_p \sum_{h \in V^{n-1}} \tilde{p}&(h) \sum_{w \in V} p(w | h) \log \frac{p(w | h)}{p_{out}(w | h)}, \label{cond-kl}
\end{align}
% \vspace{-0.03in}
while satisfying the constraints:
% \vspace{-0.03in}
\begin{align}
    \hspace{-.2in} \sum_{h \in V^{n-1}} \tilde{p}&(h) \sum_{w \in V} p(w | h) f_i(h, w) = \tilde{p}(S_i), \; i=1, \ldots, K. \label{constraints}
\end{align}
% \vspace{-0.03in}
where $f_i$ are indicator functions of $(h, w) \in S_i$. Note that Eq. \ref{cond-kl} can also be viewed as the KL divergence between the joint distribution $p(h, w)$ and $p_{out}(h, w)$ assuming they have the same history distribution $\tilde{p}(h)$. Also notice in the case that $p_{out}$ is the uniform distribution, $p_{ad}$ is indeed a maximum entropy model.

If the constraints in Equation $\ref{constraints}$ are consistent, the solution of the above optimization problem exists and is unique \cite{darroch1972generalized}. It has the following form, with parameters \{$\lambda_i$\}.
% \vspace{-0.03in}
\begin{equation}
    \hspace{-0.2in} p_{ad}(w|h) = \frac{p_{out}(w|h) * \alpha(h, w)}{Z(h, \lambda_1, \ldots, \lambda_K)}, \label{p-ad}
\end{equation}
% \vspace{-0.03in}
where the scaling factor $\alpha(h, w) = \exp{(\sum_{i=1}^{K} \lambda_i f_i(h, w))}$, and $Z(h, \lambda_1, \ldots, \lambda_K)$ is the normalization term summing up the numerators. This solution can be obtained by generalized iterative scaling (GIS) algorithm \cite{darroch1972generalized}, sketched in Algorithm \ref{algo}, or some of its modern fast counterparts \cite{Alume2010EfficientEO}. The iterations can be terminated when the results converge or nearly converge.

\setlength{\textfloatsep}{10pt}

\begin{algorithm}[t!]
\caption{Generalized Iterative Scaling (GIS) Algorithm}
\begin{algorithmic}[1]
\Require $V$, $p_{out}(w|h)$, $\tilde{p}(h)$ and $\tilde{p}(S_i)$
\State Set $\lambda_1^{(0)} = \lambda_2^{(0)} = \ldots = \lambda_K^{(0)} = 0, n = 0$
% \State n = 0
\While{stopping criterion not met}
    \State Compute $\alpha^{(n)}(h,w) = \exp\left(\sum_{i=1}^K\lambda_i^{(n)}f_i(h,w)\right)$
    \For{each seen history $h$ in training data}
        \State Compute normalization term:
        \Statex[1.7] $Z(h, \lambda_1^{(n)}, \ldots, \lambda_K^{(n)}) \coloneqq \sum_{w} p_{out}(w|h) \alpha^{(n)}(h,w)$
    \EndFor
    \State Update each entry of back-off LM:
    \Statex[1.5] $p^{(n)}(w|h) \coloneqq \frac{p_{out}(w|h) * \alpha^{(n)}(h,w)}{Z(h, \lambda_1^{(n)}, \ldots, \lambda_K^{(n)})}$
    \For{$j=1, \cdots, K$}
        \State Marginalize:
        \Statex[2.2] $p^{(n)}(S_j) \coloneqq \sum_{h} \tilde{p}(h) \sum_{w} p^{(n)}(w | h) f_j(h, w)$ 
        % E_{p_{ad}^{(n)}(w_1^3)}[f_j(w_1^3)]$ 
        % \State $\Delta_j \coloneqq \log \frac{\alpha_j}{\hat{\alpha}_j}$
        \State Update params: $\lambda_j^{(n+1)} \coloneqq \lambda_j^{(n)} + \log \frac{\tilde{p}(S_j)}{p^{(n)}(S_j)}$
    \EndFor
    \State $n \coloneqq n + 1$
\EndWhile
\State\Return $p^{(n)}(w|h)$ as $p_{ad}$
\end{algorithmic} \label{algo}
\end{algorithm}

\vspace{-0.1in}
\section{Efficient Algorithm: The Hierarchical Training Method} \label{sec-algo}

The challenge for implementing the above GIS algorithm is its computational complexity resulting from Line $4$ (normalization) and $8$ (marginalization). A naive implementation may take $O(K * \text{\# of seen histories} * |V|)$ time per iteration \cite{demori1999language}. An improvement can be made to $O(\text{\# of seen histories} * |V| + K)$ if we store the constraints in Line $7$ in a hash table and accumulate the summation in Line $8$, but this complexity is still astronomical when the corpus and vocabulary $V$ is large. Thus, we need to re-organize the summation happening in Line $4$ and $8$.

\vspace{-0.1in}
\subsection{The Hierarchical Training Method: MDI v.s. ME}

To overcome this challenge, the hierarchical training method \cite{wu2000efficient} has been proposed for ME models. The algorithm only requires linear time to its inputs per iteration, {\em i.e.}, the number of seen entries in $p_{out}$ plus $K$. The trick is based on the back-off structure of probability $p^{(n)}(w|h)$. In this paper, we show that similar algorithmic trick can be applied to MDI with additional cares. This means MDI adaptation incurs no extra computation complexity although it is more general than ME. The key is to handle the non-uniform $p_{out}$ appropriately.

% as will be shown later, this algorithm can be implemented in linear time to its inputs. The algorithmic trick is based on the backoff structure of probability $q^{(n)}(w_3|w_1^2)$, proposed in \cite{wu2000efficient}. We can share the computation of lower order $n$-grams and incrementally compute the quantities of higher order $n$-grams, whose seen $n$-grams are more and more sparse compared to the space $V^n$ and we only need to iterate through those seen $n$-grams.

% Maximum entropy (ME) models can be view as a special case of MDI adaptation where the out-domain distribution $p_{out}$ is the uniform distribution. The hierarchical training method \cite{wu2000efficient} has been proposed for ME models, and now we extend it to MDI adaptation without incurring extra computation complexity.

To illustrate the idea, we take trigram LM as an example. Consider a general LM $p(w_3|w_1^2)$ that has the back-off structure as in Equation \ref{backoff}. We also define a set of real-valued scaling factors $c(w_1^3)$, $w_1^3 \in V^3$ to be some default constant value except for $K$ of them having non-default values. Now, we hope to compute the {\em left-aligned} and {\em right-aligned} summation of the product of $p(w_3|w_1^2)$ and the general scaling factor $c(w_1^3)$:
\vspace{-0.03in}
\begin{align}
    \Sigma_L(w_1^2) &= \sum_{w_3 \in V} p(w_3|w_1^2) \cdot c(w_1^3) \label{sum1} \\
    \vspace{-0.01in}
    \Sigma_R(w_2^3) &= \sum_{w_1 \in V} p(w_3|w_1^2) \cdot c(w_1^3) \label{sum2}
\end{align}
\vspace{-0.03in}
The right-hand-sides only differ in the subscript of summation. In fact, $\Sigma_L$ corresponds to computing the normalization term (Line $4$), and $\Sigma_R$ is related to marginalization (Line $8$). Notice that in ME, $p(w_3|w_1^2)$ is just a uniform distribution.

% So, the normalization term $Z$ in line 11 is in fact a sort of $\Sigma_L$, and the $\hat{\alpha}_j$ in line 6 corresponds to $\Sigma_R$ when the feature functions $f_i$ corresponds to marginal distribution. We will show that computing the two summations above takes only linear time with regards to the number of entries in the ARPA format representing $p(w_3|w_1^2)$, {\em i.e.}, the seen $n$-grams, plus the number $K$ of features. After that, we will prove that $p_{ad}$, or $q^{(n)}(w_3|w_1^2)$, has the back-off structure as in equation \ref{backoff}, and thus the proposed algorithm can be readily applied to it.

\vspace{-0.1in}
\subsection{Computing $\Sigma_L(w_1^2)$ as normalization for history $w_1^2$} \label{sec_sum1} 
\vspace{-0.03in}

In Equation \ref{sum1}, we compute $\Sigma_L(w_1^2)$ by summing over $w_3 \in V$ given history $w_1^2$, which costs $|V|$ addition operations. However, this cost can be reduced to the number of seen $n$-grams given $w_1^2$ by dynamic programming, much less than $|V|$.

As $p(w_3|w_1^2)$ is a back-off model, we can rewrite Eq. \ref{sum1} as:
\begin{align*}
    % \hspace{-0.1in}
    \Sigma_L(w_1^2) 
    &= \;\; \sum_{w_3 \in V} p(w_3|w_1^2) \cdot c(w_1^3) \\
    &= \hspace{-0.1in} \sum_{w_3 \in V \wedge \textup{seen}(w_1^3)} \hspace{-0.2in} \left( p^*(w_3|w_1^2) - bow(w_1^2) \cdot p(w_3|w_2) \right) \cdot c(w_1^3) \\
    &+ bow(w_1^2) \sum_{w_3 \in V} p(w_3|w_2) \cdot c(w_2^3) \\
    &+ bow(w_1^2) \hspace{-0.25in} \sum_{w_3 \in V \wedge c(w_1^3) \neq c(w_2^3)} \hspace{-0.2in} p(w_3|w_2) \cdot \left( c(w_1^3) - c(w_2^3) \right).
\end{align*}
We can define $\Sigma_L(w_2) = \sum_{w_3 \in V} p(w_3|w_2) \cdot c(w_2^3)$ for the second term in the summation above. So, this appears to be a dynamic programming problem \cite{federico1999efficient, heafield2016normalized} with the base case $\Sigma_L(\emptyset) = \sum_{w_3 \in V} p(w_3) \cdot c(w_3)$, needed to compute only once. Thus, $\Sigma_L(w_1^2)$ can be computed hierarchically and bottom-up from $\Sigma_L(\emptyset)$ along the back-off structure of $p(w_3|w_1^2)$. 

As of complexity, computing $\Sigma_L(\emptyset)$ requires $O(|V|)$ time. $\Sigma_L(w_1^2)$ and $\Sigma_L(w_2)$ can be computed from the $\Sigma_L$ of the lower-order $n$-grams with the complexity of the number of seen $n$-grams along the way, plus the number of distinct scaling factors $K$ (at most), as each non-default $c(\cdot)$ term is accessed only once. Thus, overall, the total time is proportional to the number of seen entries in $p_{out}(w_3|w_1^2)$ plus K. Recall in the ME case, $p(\cdot)$ and $bow(\cdot)$ can be seen as $1$, and the above equation can be simplified to contain only the scaling factor $c(\cdot)$'s \cite{wu2000efficient}.

% same principle as ME, log-linear, efficient_mdi

\subsection{Computing $\Sigma_R$}

Consider Eq. \ref{sum2}, where we compute the right-aligned $\Sigma_R(w_2^3)$ by summing over $w_1 \in V$. Similarly, we define the lower or higher order right-aligned summation $\Sigma_R(w_3)$ as follows:
\begin{align}
    \Sigma_R(w_3) &= \sum_{w_1^2 \in V^2} p(w_3|w_1^2) \cdot c(w_1^3) \label{sum2} \\
    \Sigma_R(w_1^3) &= p(w_3|w_1^2) \cdot c(w_1^3)
\end{align}
Obviously, $\Sigma_R(w_3)$ requires more computation than $\Sigma_R(w_2^3)$ and $\Sigma_R(w_1^3)$. Unfortunately, dynamic programming does not work here anymore to compute $\Sigma_R(w_3)$ from $\Sigma_R(w_2^3)$. Instead, we will make use of the idea of {\em shared computation}, which means we go over the data for only one pass, but we accumulate the values correspondingly to all related constraints.

First, let us compute $\Sigma_R(w_2^3)$ as follows:
\begin{align*}
    % \hspace{-0.1in}
    \Sigma_R(w_2^3) 
    &= \;\; \sum_{w_1 \in V} p(w_3|w_1^2) \cdot c(w_1^3) \\
    &= \hspace{-0.15in} \sum_{w_1 \in V \wedge \textup{seen}(w_1^3)} \hspace{-0.2in} \left( p^*(w_3|w_1^2) - bow(w_1^2) \cdot p(w_3|w_2) \right) \cdot c(w_1^3) \\
    &+ \quad p(w_3|w_2) \cdot c(w_2^3) \sum_{w_1 \in V} bow(w_1^2) \\
    &+ \hspace{-0.25in} \sum_{w_1 \in V \wedge c(w_1^3) \neq c(w_2^3)} \hspace{-0.2in} p(w_3|w_2) \cdot bow(w_1^2) \cdot \left( c(w_1^3) - c(w_2^3) \right)  
\end{align*}
For simplicity, we denote the auxiliary function $g(w_2) = \sum_{w_1 \in V} bow(w_1^2)$ in the second term above, and we will address the computation of $g(\cdot)$ later. Notice that the decomposition of $\Sigma_R$ looks quite different from section \ref{sec_sum1}, as $\Sigma_R(w_2^3)$ is not decomposed to the sub-problem $\Sigma_R(w_3)$ or vice versa. Besides, if we let $p(\cdot)$ and $bow(\cdot)$ be $1$, it becomes the case for ME model.

At the same time, let us see how to compute $\Sigma_R(w_3)$:
% \begin{align*}
%     \hspace{-0.1in}
%     \Sigma_R(w_1^3)
%     &= p(w_3|w_1^2) \cdot c(w_1^3)
% \end{align*}
%
% \begin{align*}
%     \hspace{-0.1in}
%     \Sigma_R(w_3)
%     &= \sum_{w_1^2 \in V^2} p(w_3|w_1^2) \cdot c(w_1^3) \\
%     %
%     &= \quad \sum_{w_2 \in V} \; \Sigma_R(w_2^3) \\
%     %
%     &= \sum_{\substack{w_1^2 \in V^2 \\ \wedge \textup{seen}(w_1^3)}} \left( p^*(w_3|w_1^2) - bow(w_1^2) \cdot p(w_3|w_2) \right) \cdot c(w_1^3) \\
%     %
%     &+ \quad \sum_{w_2 \in V} \; p(w_3|w_2) \cdot c(w_2^3) \cdot g(w_2) \\
%     %
%     &+ \sum_{\substack{w_1^2 \in V^2 \\ \wedge c(w_1^3) \neq c(w_2^3)}} p(w_3|w_2) \cdot bow(w_1^2) \cdot \left( c(w_1^3) - c(w_2^3) \right) \\
%     %
%     &= \sum_{\substack{w_1^2 \in V^2 \\ \wedge \textup{seen}(w_1^3)}} \left( p^*(w_3|w_1^2) - bow(w_1^2) \cdot p(w_3|w_2) \right) \cdot c(w_1^3) \\
%     %
%     &+ \sum_{\substack{w_2 \in V \\ \wedge \textup{seen}(w_2^3)}} \left( p^*(w_3|w_2) - bow(w_2) \cdot p(w_3) \right) \cdot c(w_2^3) \cdot g(w_2) \\
%     %
%     &+ \quad p(w_3) \cdot c(w_3) \sum_{w_2 \in V} \;  bow(w_2) \cdot g(w_2) \\
%     %
%     &+ \sum_{\substack{w_2 \in V \\ \wedge c(w_2^3) \neq c(w_3)}} p(w_3) \cdot bow(w_2) \cdot \left( c(w_2^3) - c(w_3) \right) \\
%     %
%     &+ \sum_{\substack{w_1^2 \in V^2 \\ \wedge c(w_1^3) \neq c(w_2^3)}} p(w_3|w_2) \cdot bow(w_1^2) \cdot \left( c(w_1^3) - c(w_2^3) \right)
% \end{align*}
%
\begin{align*}
    \hspace{-0.1in}
    \Sigma_R(w_3)
    % &= \sum_{w_1^2 \in V^2} p(w_3|w_1^2) \cdot c(w_1^3) \\
    % %
    % &= \quad \sum_{w_2 \in V} \; \Sigma_R(w_2^3) \\
    % %
    % &= \sum_{\substack{w_1^2 \in V^2 \\ \wedge \textup{seen}(w_1^3)}} \left( p^*(w_3|w_1^2) - bow(w_1^2) \cdot p(w_3|w_2) \right) \cdot c(w_1^3) \\
    % %
    % &+ \quad \sum_{w_2 \in V} \; p(w_3|w_2) \cdot c(w_2^3) \cdot g(w_2) \\
    % %
    % &+ \sum_{\substack{w_1^2 \in V^2 \\ \wedge c(w_1^3) \neq c(w_2^3)}} p(w_3|w_2) \cdot bow(w_1^2) \cdot \left( c(w_1^3) - c(w_2^3) \right) \\
    % %
    &= \hspace{-0.1in} \sum_{w_1^2 \in V^2 \wedge \textup{seen}(w_1^3)} \hspace{-0.2in} \left( p^*(w_3|w_1^2) - bow(w_1^2) \cdot p(w_3|w_2) \right) \cdot c(w_1^3) \\
    &+ \hspace{-0.1in} \sum_{w_2 \in V \wedge \textup{seen}(w_2^3)} \hspace{-0.2in} \left( p^*(w_3|w_2) - bow(w_2) \cdot p(w_3) \right) \cdot c(w_2^3) \cdot g(w_2) \\
    &+ \quad p(w_3) \cdot c(w_3) \sum_{w_2 \in V} \;  bow(w_2) \cdot g(w_2) \\
    &+ \hspace{-0.2in} \sum_{w_2 \in V \wedge c(w_2^3) \neq c(w_3)} \hspace{-0.25in} p(w_3) \cdot bow(w_2) \cdot \left( c(w_2^3) - c(w_3) \right) \\
    &+ \hspace{-0.2in} \sum_{w_1^2 \in V^2 \wedge c(w_1^3) \neq c(w_2^3)} \hspace{-0.25in} p(w_3|w_2) \cdot bow(w_1^2) \cdot \left( c(w_1^3) - c(w_2^3) \right)
\end{align*}
We denote $g(\emptyset) = \sum_{w_2 \in V} bow(w_2) \cdot g(w_2)$. Assuming that the values of $g(\cdot)$ are known, now we can come up with an algorithm to compute $\Sigma_R$ by observing the equations of $\Sigma_R(w_3)$, $\Sigma_R(w_2^3)$ and $\Sigma_R(w_1^3)$ together. The algorithm enumerates the seen $n$-grams in LM $p(w_3|w_1^2)$ and all non-default scaling factor $c(\cdot)$'s, and adds the value as specified in the equations to the $\Sigma_R$ with arguments matching the suffix of the $n$-gram. The complexity is $O(n * \text{\# of entries in $p(w|h)$})$, with $n$ being a small constant.

Now, the remaining problem is how to compute the auxiliary function $g(\cdot)$ as defined previously. It turns out that this is a right-aligned summation in the ME case. More specifically, their scaling factors are $c(w_1^2) = bow(w_1^2)$ or $c(w_1^2) = bow(w_1^2)*bow(w_2)$. Thus, computing $g(\cdot)$ can be shown to be also in linear time. In fact, computing the auxiliary function $g(\cdot)$ is what makes the algorithm for MDI different from that of ME.
% is the so-called ``shared computation'' between $n$-grams with the same suffix, {\em and exists solely for MDI compared to ME}. \textcolor{red}{ML approximation of p(h)}
% 1. shared computation
% 2. history can be smoothed

% \begin{align*}
%     \Sigma_R(w_3) 
%     &= \quad \Sigma_R(w_2^3) \\
%     %
%     &- \quad p(w_3|w_2) \cdot c(w_2^3) \cdot g(w_2) \\
%     %
%     &+ \sum_{\substack{w_2 \in V \\ \wedge \textup{seen}(w_2^3)}} \left( p^*(w_3|w_2) - bow(w_2) \cdot p(w_3) \right) \cdot c(w_2^3) \cdot g(w_2) \\
%     %
%     &+ \quad p(w_3) \cdot c(w_3) \cdot g(\emptyset) \\
%     %
%     &+ \sum_{\substack{w_2 \in V \\ \wedge c(w_2^3) \neq c(w_3)}} p(w_3) \cdot bow(w_2) \cdot \left( c(w_2^3) - c(w_3) \right)
% \end{align*}
% That is, $\Sigma_R(w_3)$ can be incrementally computed from $\Sigma_R(w_2^3)$ in a top-down manner, in contrast to computing $\Sigma_L$.

\subsection{The back-off structure of $p_{ad}(w|h)$}

Before computing the marginals in Line 8 of Algorithm \ref{algo}, we still need to show that the probability $p^{(n)}(w|h)$ or $p_{ad}(w|h)$ in Equation \ref{p-ad} has the back-off structure. This is important not only for the computational purpose -- that the tricks for $\Sigma_L$ and $\Sigma_R$ can be applied here -- but also for being able to represent the final adapted LM in ARPA format.

% The adapted model can be written as:
% \begin{align*}
%     p_{ad}(w_3|w_1^2) = \frac{p_{out}(w_3|w_1^2) \cdot c(w_1^3)}{Z(w_1^2)}
% \end{align*}
We claim that, if $p_{out}$ is a back-off model as in Equation \ref{backoff}, then so is the exponential models $p^{(n)}$ and $p_{ad}$. We prove this by giving the back-off expression of $p_{ad}$:
\begin{equation*}
      p_{ad}(w_3|w_1^2) = \left\{
        \begin{array}{ll}
            p_{ad}^*(w_3|w_1^2) & \textup{if } w_1^3 \textup{ seen in } p_{out} \\
            & \textup{or } w_1^3 \textup{ is a constraint} \\
            bow_{ad}(w_1^2) \cdot p_{ad}(w_3|w_2)   & \textup{otherwise} \\
		\end{array} 
		\right.
\end{equation*}
\vspace{-0.1in}
where:
\begin{align*}
    p_{ad}^*(w_3|w_1^2) &= \frac{p_{out}(w_3|w_1^2) \cdot c(w_1^3)}{Z(w_1^2)} \\
    bow_{ad}^*(w_1^2) &= \frac{Z(w_2)}{Z(w_1^2)} bow_{out}(w_1^2)
\end{align*}
The lower order $n$-grams of $p_{ad}$ are defined analogously. There will be at most $(\textup{\# entries in } p_{out} + \textup{\# entries in } p_{in})$ entries in $p_{ad}$, same as in linear interpolation.

\subsection{Computing marginalization as $\Sigma_R$}

Finally, we come to compute the marginals in Line 8 of Algo. \ref{algo}:
\begin{align}
    p^{(n)}(S_i) \coloneqq \sum_{h} \tilde{p}(h) \sum_{w} p^{(n)}(w | h) f_i(h, w).
\end{align}
Since it has been proved that $p^{(n)}$ is a back-off LM, we can view $\sum_{w} p^{(n)}(w | h) f_i(h, w)$ as a right-aligned sum. However, we need to further consider the multiplication term $\tilde{p}(h)$. Fortunately, the same trick computing $\Sigma_R$ can be applied here, with some modification of the auxiliary function $g$. For example, let $g(w_2) = \sum_{w_1 \in V} \tilde{p}(w_1^2) bow(w_1^2)$, and then it can be treated in two ways efficiently, either (i) if $\tilde{p}(w_1^2)$ is an unsmoothed maximum likelihood estimation, there will be a lot of zeros for $\tilde{p}(w_1^2)$, or (ii) if $\tilde{p}(w_1^2) = \tilde{p}(w_2 | w_1) * \tilde{p}(w_1)$ has a smoothed distribution, and $\tilde{p}(w_2 | w_1)$ has the back-off structure, then this amounts to compute some right-aligned sum. It can be shown that the computational complexity of marginalization is linear in both ways. We omit the details here due to space limit. Interested readers can refer to Appendix A at the end of the paper. %to Appendix A. % \cite{arXivVersion}
In all, we have shown how Algorithm \ref{algo} can be implemented efficiently.

\subsection{Implementation Issues}

Special care should be taken when dealing with $n$-grams $w_1^3$ which containing \bos or \eos, or whose suffix $w_2^3$ is not seen. To further speed up the computation, the algorithm can be implemented in a vectorized manner with group-by operation for summing up probabilities of $n$-grams of the same suffix.

% (1) be cautious handling ``$<s>$'' and ``$</s>$'';
% (2) when $p_{out}$ is given as conditional probability, we need to  approximately recover the joint probability, i.e., we need to come up with $p_h$ and compute the marginal probabilities for the arpa LM;
% (3) magnitude of lambdas, overflow, constraint selection, constraint interaction; (4) the choice of vocabulary; (5) vectorization

\section{Experimental Results}

% \begin{center}
\begin{table*}[!htb]
% \hspace*{-0.6cm}
\centering
\tabcolsep=0.17cm
\begin{tabular}{cc|cccc|ccccccccc}
\hline
\multirow{3}{*}{\textbf{Corpus}} & \multirow{3}{*}{\textbf{Test set}} & \multicolumn{4}{c|}{\textbf{First-pass LM}}                 & \multicolumn{9}{c}{\textbf{Rescoring with large $n$-gram LM}}                                                                                                                                 \\ \cline{3-15} 
                                 &                                    & \multicolumn{2}{c|}{default}     & \multicolumn{2}{c|}{MDI} & \multicolumn{1}{c|}{No adapt.} & \multicolumn{2}{c|}{Interpolation}                   & \multicolumn{2}{c|}{MDI (2-2-2)} & \multicolumn{2}{c|}{MDI (5-3-2)} & \multicolumn{2}{c}{MDI (6-4-3)} \\
                                 &                                    & PPL   & \multicolumn{1}{c|}{WER} & PPL    & WER             & \multicolumn{1}{c|}{PPL}       & PPL                       & \multicolumn{1}{c|}{WER} & PPL   & \multicolumn{1}{c|}{WER} & PPL   & \multicolumn{1}{c|}{WER} & PPL        & WER                \\ \hline
\multirow{2}{*}{AMI}             & dev                                & 84.6  & 20.0                     & 84.3   & 20.0            & 384.1                          & 80.5                      & 19.6                     & 86.6  & 19.4                     & 87.1  & 19.4                     & 87.9       & \textbf{19.4}      \\
                                 & eval                               & 79.7  & 20.2                     & 79.9   & 20.2            & 408.8                          & 77.5                      & 20.0                     & 81.8  & 19.6                     & 82.8  & 19.6                     & 83.9       & \textbf{19.6}      \\ \hline
\multirow{3}{*}{SWBD}            & dev                                & 98.6  & 12.5                     & 96.9   & \textbf{12.0}  & 411.0                          & 92.7                      & \textbf{11.4}           & 94.5  & 11.7                    & 95.1  & 11.7                    & 95.8       & 11.6              \\
                                 & eval2000                           & 179.2 & 14.2                     & 117.5  & \textbf{14.0}   & 161.6                          & 85.6                      & 13.4                     & 88.9  & 13.2                     & 89.4  & 13.2                     & 89.4       & \textbf{13.2}      \\
                                 & rt03                               & 167.8 & 17.3                     & 109    & \textbf{17.2}   & 149.4                          & 78.6                      & 16.3                     & 82.2  & 16.2                     & 82.7  & 16.1                     & 82.7       & \textbf{16.1}      \\ \hline
\multirow{2}{*}{WSJ}             & dev93                              & 186.6 & 7.0                     & 161.3  & \textbf{6.8}   & 223.3                          & \multicolumn{1}{l}{134.2} & 6.3                     & 134.8 & \textbf{6.2}            & 135.1 & 6.3                     & 136.8      & 6.3               \\
                                 & eval92                             & 164.8 & 4.7                     & 142.7  & 4.7   & 222.2                          & \multicolumn{1}{l}{118.7} & 4.0                      & 117.4 & 3.9                     & 118.0 & \textbf{3.9}            & 120.0      & 3.9               \\ \hline
\end{tabular}
\caption{Comparing the perplexity (PPL) and word error rate (WER, in \%) of LMs with no adaptation, interpolation and MDI adaptation.}
\vspace{-.25in}
\label{table1}
\end{table*}
% \end{center}

% unbiased rounding-up rule:
% >0.5 
% <0.5 
% =0.5: make even number

We will show the scalability of our algorithm and the effectiveness of MDI adaptation with two different ways of application. 

\pagebreak

% \textcolor{red}{four}

We simulate the LM adaptation scenario by taking three speech corpora, Wall Street Journal (WSJ), Switchboard (SWBD) and AMI-IHM (AMI) as in-domain data, and Google One Billion Words \cite{chelba2013one} and Librispeech \cite{panayotov2015librispeech} as out-of-domain data. We normalize the Google dataset with the similar scripts generating normalized Librispeech LM training texts, resulting $702$ million and $803$ million words respectively. We compute trigram LMs using the SRILM tool \cite{Stolcke2002SRILMA} with Kneser-Ney smoothing and default settings, or use default Kaldi's LMs. We find the normalized Google dataset always out-perform Librispeech as the out-of-domain corpus, so we only report the results for Google dataset. We are interested whether the rich LM information in the very large corpus can help the LM and ASR task in the application domains. We use count thresholds to select the in-domain constraints, {\em i.e.}, the marginals are considered reliable when the counts of the $n$-grams is above the threshold.

\begin{figure}[t]
  \centering
  \includegraphics[width=\linewidth]{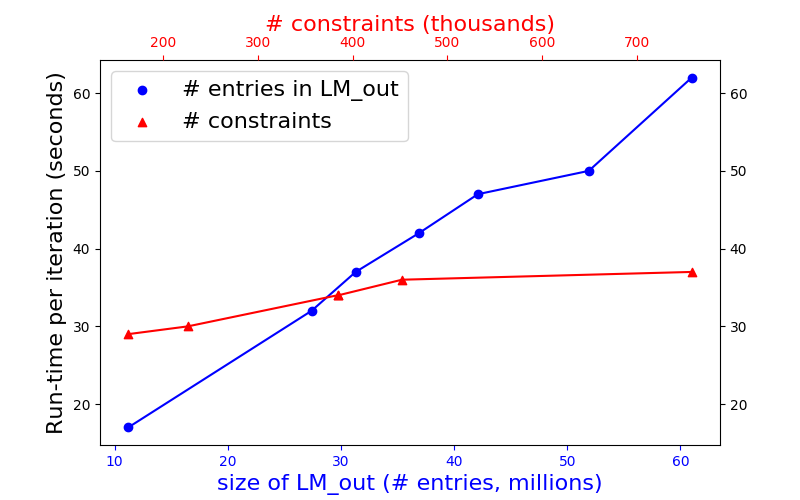}
  \vspace{-0.25in}
  \caption{Run-time of Algorithm 1 with various input sizes.}
  \vspace{-0.05in}
  \label{fig_scalability}
  \vspace{-0.05in}
\end{figure}

\vspace{-0.05in}
\subsection{Scalability}
% \vspace{-0.05in}

We implemented the proposed efficient version of Algorithm \ref{algo} in Python with Numpy and Pandas. In Figure \ref{fig_scalability}, we compare the run-time per iteration in seconds with various size of out-of-domain LM (blue) and various number of constraints (blue). The in- and out-of-domain data are taken to be SWBD and Google. We sample the out-of-domain data at different sizes measured by the total number of seen entries in the ARPA file, and record the run-time per iteration. We control the number of constraints by using different constraint count thresholds. We can see that both lines shows linear scalability, and the run-time is denominated by the size of out-of-domain data. Besides, it usually takes $60 \sim 80$ iterations for the algorithm to converge to a near optimal solution, which may be improved by more advanced optimization algorithms.

\vspace{-0.05in}
\subsection{Effectiveness of Adaptation}
% \vspace{-0.05in}

We compare the LMs with and without adaptation, and with different adaptation methods, {\em i.e.}, simple linear interpolation and MDI. We evaluate the LMs in perplexity (PPL) and word error rate (WER) when used in an hybrid ASR system. We use the latest recipes in the open-source speech recognition toolkit Kaldi \cite{povey2011kaldi} to run the ASR experiments.
The acoustic model uses factorized TDNN architecture \cite{povey2018semi} and is trained with LF-MMI criteria \cite{povey2016purely}. The features are 40-dimensional MFCC features with 100-dimensional i-vectors appended to the MFCC. The training data is augmented with speed and volume perturbation.

% We compare the LMs with and without adaptation, and with different adaptation methods, {\em i.e.}, simple linear interpolation and MDI. We evaluate the LMs in perplexity and word error rate (WER) when used in an hybrid ASR system. We use the open-source speech recognition toolkit Kaldi \cite{povey2011kaldi} to run the ASR experiments. We evaluate the algorithm for ASR tasks on four speech corpora, Wall Street Journal (WSJ), Switchboard (SWBD), AMI-IHM, and Tedlium. For neural network training, we use the HMM-GMM system as the seed model to get alignments from the training data. The DNN is trained on 40-dimensional MFCC features, and 100-dimensional i-vectors are appended to the MFCC. For the neural network, we use factorized TDNN architecture \cite{povey2018semi} and train the acoustic model with LF-MMI training criteria \cite{povey2016purely}. To improve model robustness, we augment the training data with speed perturbation and volume perturbation.

% To improve model robustness, we augment the training data with speed perturbation and volume perturbation.

% \goodbreak

\vspace{-0.05in}
\subsubsection{Performance in the Rescoring}

As in a common adaptation scenario, we adapt the large, out-of-domain LM to satisfy the marginal distribution constraints derived from the in-domain data. As the Google corpus is large, the resulting LM can have $61 \sim 76$ million entries depending on the vocabulary (which is AMI $50$k, SWBD $30$k, WSJ $20$k). So the LMs are used for rescoring. From the right half of Table \ref{table1}, we can see that LM adaptation is effective for both interpolation and MDI. The interpolated LMs have better perplexity most of the time, which is consistent with previous work \cite{rao1997mdi, chen2009shrinking}, but we also find that MDI adapted LMs have better WER. Also note the constraints we use for MDI, where $5$-$3$-$2$ means count thresholds of $5, 3, 2$ are used for selecting unigrams, bigrams and trigrams as constraints. Thus, in fact, MDI sees less information of the in-domain data than interpolation, {\em e.g.}, the MDI($6$-$4$-$3$) above sees the least in-domain data. However, MDI requires additional information about the history distribution $\tilde{p}(h)$, for which we take the maximum-likelihood in-domain $p_{in}(h)$ as an approximation.

\vspace{-0.05in}
\subsubsection{Performance in the First-Pass Decoding}

We propose a novel approach of applying MDI adaptation: instead of adapting the out-of-domain LM, we adapt the small, in-domain LM, which is used in the first pass decoding of ASR, so that it preserves the marginals of the larger and better interpolated LM. The constraints are selected to be all seen entries in the in-domain LM, such that the model size remains unchanged after adaptation. The results are a bit surprising. As we can see in the left half of Table \ref{table1}, the perplexity of the first-pass LM gets improved significantly and lies between the default and interpolated LMs. The first-pass WER also gets improved, not so much though. It seems this is the advantage of MDI over interpolation: we have better control of the resulting model size. We are going to investigate into this interesting observation as the future work.

% something regarding language model specific to dataset
% We use a 3-gram LM trained with the SRILM
% toolkit in the first pass decoding. We
% perform a lattice rescoring.

% \begin{table}[t]
% \centering
% \caption{Diarization results for track 2.}
% \label{tab:diar_results}
% \begin{tabular}{lcccc}
% \toprule
% \multicolumn{1}{c}{{\textbf{System}}} & \multicolumn{2}{c}{\textbf{Dev}} & \multicolumn{2}{c}{\textbf{Eval}} \\
% \multicolumn{1}{c}{}                                 & \textbf{DER}    & \textbf{JER}   & \textbf{DER}    & \textbf{JER}    \\
% \midrule
% Baseline (U06)  & 63.42  & 70.83 & 68.20 & 72.54  \\
% PLDA Fusion & 63.97 & 71.65 & 71.56 & 71.32 \\
% + 0.25s shift   & 61.00 & 66.23 & 69.64 & 69.81 \\
% + overlap assign. & 58.18 & 59.92 & 69.92 & 65.64  \\
% \bottomrule
% \end{tabular}
% \end{table}

% Settings: (1) comparing different constraints, (2) comparing different datasets, formal or conversational, (3) using the in-domain vocabulary vs. using larger-sized or out-domain vocabulary.

\vspace{-0.05in}
\section{Conclusion and Future Work}
\vspace{-0.05in}

In this paper, we propose an efficient MDI adaptation algorithm for $n$-gram LMs. The algorithm relies on the back-off structure of the LMs, and takes linear time per iteration. We show empirically our algorithm is truly scalable to very large corpus. We also find that MDI adaptation gets close perplexity to linear interpolation, but better WER. The methods for computing marginals and normalization terms are general and may benefit some more advanced optimization algorithms. Regarding MDI models, it may be important to study whether the better feature selection and history distribution estimation methods can affect the performance. Lastly, as we have observed in the experiments, we will study using MDI adaptation to improve small LMs for the first-pass decoding of ASR.

% 1. feature selection, history estimation;
% 2. implement regularization in GIS;
% 3. KN consistency.

% \section{Acknowledgements}

% The authors would like to thank Matthew Wiesner for his valuable suggestions about the optimization algorithms.

\newpage

\onecolumn

\section{Appendix A}

\twocolumn
\bibliographystyle{IEEEtran}

\bibliography{mybib}

% \begin{thebibliography}{9}
% \bibitem[1]{Davis80-COP}
%   S.\ B.\ Davis and P.\ Mermelstein,
%   ``Comparison of parametric representation for monosyllabic word recognition in continuously spoken sentences,''
%   \textit{IEEE Transactions on Acoustics, Speech and Signal Processing}, vol.~28, no.~4, pp.~357--366, 1980.
% \bibitem[2]{Rabiner89-ATO}
%   L.\ R.\ Rabiner,
%   ``A tutorial on hidden Markov models and selected applications in speech recognition,''
%   \textit{Proceedings of the IEEE}, vol.~77, no.~2, pp.~257-286, 1989.
% \bibitem[3]{Hastie09-TEO}
%   T.\ Hastie, R.\ Tibshirani, and J.\ Friedman,
%   \textit{The Elements of Statistical Learning -- Data Mining, Inference, and Prediction}.
%   New York: Springer, 2009.
% \bibitem[4]{YourName17-XXX}
%   F.\ Lastname1, F.\ Lastname2, and F.\ Lastname3,
%   ``Title of your INTERSPEECH 2020 publication,''
%   in \textit{Interspeech 2020 -- 20\textsuperscript{th} Annual Conference of the International Speech Communication Association, September 15-19, Graz, Austria, Proceedings, Proceedings}, 2020, pp.~100--104.
% \end{thebibliography}

\end{document}